\documentclass{article}

\usepackage[preprint]{neurips_2025}

\usepackage[utf8]{inputenc}    
\usepackage[T1]{fontenc}       

\usepackage{microtype}         
\usepackage{booktabs}          
\usepackage{nicefrac}          

\usepackage{amsmath, amsthm, amsfonts}
\newtheorem{theorem}{Theorem}
\newtheorem{lemma}{Lemma}

\usepackage{xcolor}            

\usepackage{graphicx}
\usepackage{subcaption}

\usepackage{hyperref}
\usepackage{url}

\usepackage{listings}
\title{Rethinking RoPE: A Mathematical Blueprint for N-dimensional Rotary Positional Embedding}

\author{%
  \textbf{Haiping Liu} \quad
  \textbf{Lijing Lin} \quad
  \textbf{Jingyuan Sun} \quad
  \textbf{Zhegong Shangguan} \\
  \textbf{Mauricio A. Alvarez} \quad
  \textbf{Hongpeng Zhou}$^{*}$ \\
  University of Manchester \\
  \texttt{* Corresponding author: hongpeng.zhou@manchester.ac.uk}
}

\begin{document}

\maketitle

\begin{abstract}
Rotary Position Embedding (RoPE) is widely adopted in large language models (LLMs) due to its efficient encoding of relative positions with strong extrapolation capabilities. 
However, while its application 
in higher-dimensional input domains, such as 2D images, have been explored in several attempts, a unified theoretical framework is still lacking.
To address this, we propose a systematic mathematical framework for RoPE grounded in Lie group and Lie algebra theory. 
We derive the necessary and sufficient conditions for any valid $N$-dimensional RoPE based on two core properties of RoPE - relativity and reversibility.
We demonstrate that RoPE can be characterized as a basis of a maximal abelian subalgebra (MASA) in the special orthogonal Lie algebra, and that the commonly used axis-aligned block-diagonal RoPE, where each input axis is encoded by an independent 2×2 rotation block, corresponds to the maximal toral subalgebra.
Furthermore, we reduce spatial inter-dimensional interactions to a change of basis, resolved by learning an orthogonal transformation.
Our experiment results suggest that inter-dimensional interactions should be balanced with local structure preservation.
Overall, our framework unifies and explains existing RoPE designs while enabling principled extensions to higher-dimensional modalities and tasks.
\end{abstract}

\vspace{-1em}
\section{Introduction}
\label{sec:introduction}
Rotary Position Embedding (RoPE)~\citep{kexue, roformer} has become a standard positional embedding method in Transformer-based \citep{transformer} large language models (LLMs) \citep{llama3, deepseek}, for its parameter-free efficiency, seamless integration with attention mechanisms, and robust extrapolation capabilities to longer sequences.
RoPE has recently been extended to two-dimensional (2D) spatial inputs, attracting increasing interest in computer vision~\citep{visionllama, naver-rope, liere, fit} and multi-modal learning~\citep{Qwen, unified-io}.
However, this 2D extension introduces additional complexity, 
including how to preserve the desirable properties of the original RoPE designed for 1D language and how to systematically construct multiple rotation operations across spatial dimensions to capture their interactions within a unified structure.

As a result, some existing approaches, such as \citep{visionllama, Qwen}, construct 2D RoPE by simply duplicating the original 1D RoPE along separate axes, which we refer to as standard 2D RoPE, without exploring its underlying mathematical structure.
\citep{naver-rope, liere} propose learnable extensions to RoPE, but they remain empirical and lack theoretical grounding. 
STRING \citep{string} is the most related concurrent work, which enhances RoPE by introducing learnable parameters to model interactions between spatial dimensions. Although it proposes a similar form and demonstrates its ability to preserve relative positions, it does not identify the essential conditions for constructing a valid RoPE, nor does it derive the structural principles.
This highlights a broader issue: 
despite RoPE’s empirical success,
existing work lacks a theoretical framework for extending RoPE to higher-dimensional settings.

To address this theoretical gap, we propose a unified mathematical framework for RoPE grounded in Lie group and Lie algebra theory. Starting from two fundamental properties, \textit{relativity} and \textit{reversibility}, we derive the necessary and sufficient conditions that any valid RoPE must satisfy. We further prove that all valid \(N\)-dimensional RoPEs lie within the basis of a maximal abelian subalgebra (MASA) of the special orthogonal Lie algebra \(\mathfrak{so}(n)\), and that the standard RoPE, including both 1D and 2D forms, corresponds to the maximal toral subalgebra. Finally, we reduce the problem of modeling inter-dimensional interactions to learning an orthogonal matrix, and introduce several parameterization strategies under this formulation.
Through both empirical comparisons and theoretical analysis of different learnable parameterizations of the orthogonal matrix, we find that the Householder transformation \citep{jhu_matrix}, which better preserves local structure, tends to yield superior performance in downstream tasks. 
In contrast, the Cayley transform \citep{optimization} introduces global interactions that may disrupt spatial locality. These findings suggest that when designing learnable RoPE, one should carefully balance the preservation of localized rotational structure with the need for flexible inter-dimensional interaction.

\section{Preliminaries}
\label{sec:preliminary}

\subsection{Lie Group and Lie Algebra}
\label{subsec:Lie Group}
Lie groups are smooth manifolds with group structure, and their associated Lie algebras describe the local, infinitesimal behavior near the identity. Notably, this paper primarily focuses on the special orthogonal Lie group $\mathrm{SO}(n)$ and its associated Lie algebra $\mathfrak{so}(n)$.
Specifically, $\mathrm{SO}(n)$ is defined as the set of all real $n \times n$ orthogonal matrices with determinant equal to one, denoting as:
\[
\mathrm{SO}(n) = \left\{ A \in \mathbb{R}^{n \times n} \;\middle|\; A^T A = I,\; \det(A) = 1 \right\}.
\]
Importantly, matrices in $\mathrm{SO}(n)$ preserve both Euclidean norms and orientations, and are thus commonly referred to as rotation matrices. 
Closely related to $\mathrm{SO}(n)$ is its Lie algebra $\mathfrak{so}(n)$, which captures the infinitesimal structure of the group, consisting of all real $n \times n$ skew-symmetric matrices:
\[
\mathfrak{so}(n) = \left\{ X \in \mathbb{R}^{n \times n} \;\middle|\; X^T = -X \right\}.
\]
Furthermore, the connection between the Lie algebra $\mathfrak{so}(n)$ and the Lie group $\mathrm{SO}(n)$ is established through the matrix exponential, which offers a local diffeomorphism between the two. Specifically, for any $X \in \mathfrak{so}(n)$, the matrix exponential yields an element in $\mathrm{SO}(n)$:
\(
\exp(X) \in \mathrm{SO}(n).
\)
Additionally, a key algebraic property of the matrix exponential relevant to this paper is the following rule:
\begin{equation}
    \label{eq:}
    XY = YX \quad \Longrightarrow \quad \exp(X + Y) = \exp(X)\exp(Y) 
\end{equation}
i.e., exponential additivity holds for commuting matrices.
In the context of Lie algebras, this condition is referred to as commutativity, characterized by 
\(
[X, Y] = XY - YX = 0.
\)

\vspace{-0.5em}
\subsection{Essential Properties of RoPE}
\label{subsec: Essential Properties of RPE}
The two key properties of RoPE are \textit{relativity} and \textit{reversibility} \citep{kexue}.
\textit{Relativity} refers to the relative distance between any two tokens can be derived from their absolute positions in a way that preserves the structure of self-attention, which is a crucial factor for enabling extrapolation ability of Transformer.  
Let \( \boldsymbol{x}_1 \) and \( \boldsymbol{x}_2 \) be two absolute positions in \(N\)-dimensional Euclidean space, representing the positions of a query and a key, respectively.  
We denote the associated RoPE matrices as \( R_{\boldsymbol{x}_1}, R_{\boldsymbol{x}_2} \in \mathbb{R}^{d \times d} \).

To satisfy the \textit{relativity} requirement, $R_{\boldsymbol{x}}$ must ensure that applying RoPE at each absolute position yields an attention score identical to that obtained by applying a single transformation that depends only on their relative offset between the positions.
Formally:
\[
    (R_{\boldsymbol{x}_1} q)^T (R_{\boldsymbol{x}_2} k) = q^T R_{\boldsymbol{x}_1}^T R_{\boldsymbol{x}_2} k = q^T R_{\boldsymbol{x}_2 - \boldsymbol{x}_1} k.
\]

That is, the relative distance should be encoded through absolute positions like:
\begin{equation}
    \label{eq:nd relativity}
    R_{\boldsymbol{x}_1}^T R_{\boldsymbol{x}_2} = R_{\boldsymbol{x}_2 - \boldsymbol{x}_1}.
\end{equation}

\textit{Reversibility}, on the other hand, requires that the mapping $\boldsymbol{x} \mapsto R_{\boldsymbol{x}}$ be injective. This ensures that the absolute position can be uniquely recovered from its encoding without information loss. Formally, this condition is expressed as:
\begin{equation}
    \label{eq:nd reversibility}
    R_{\boldsymbol{x}_1} = R_{\boldsymbol{x}_2} \;\; \Rightarrow \;\; \boldsymbol{x}_1 = \boldsymbol{x}_2
\end{equation}

\subsection{The General Form of RoPE}
\label{subsec:general form of rope}
The goal of this paper is to construct a position-dependent transformation \( R_{\boldsymbol{x}} \) that satisfies both \textit{relativity} and \textit{reversibility}, as characterized by Equations~\eqref{eq:nd relativity} and~\eqref{eq:nd reversibility}.
To this end, inspired by the structure-preserving properties of the matrix exponential, we adopt an exponential parameterization of the form:
\begin{equation}
    \label{eq:rope form}
    R = \exp(\sum_{i=1}^{N} x_i B_i),
\end{equation}
where \(x_i\) denotes the spatial coordinate along the \(i\)-th axis, and \(B_i\) is the generator matrix associated with that axis.
Under this formulation, the task reduces to identifying a valid set of matrices \( \{B_i\} \) that ensures \( R \) satisfies the desired properties.

For clarity in the derivation, the matrix \(B_i\) and its corresponding RoPE matrix introduced in this paper adopt the minimal dimensionality of \(2N\), which suffices to satisfy the essential properties of RoPE for \(N\)-dimensional constructions and provides adequate freedom along spatial axes.
The detailed explanation and discussion of this dimensionality choice is deferred to Section~\ref{N-dimensional RoPE construction via MASA}.
In practice, Transformer implementations typically extend or repeat such constructions across frequency bands to match the full dimensionality of the query and key vectors.

\vspace{-0.5em}
\section{Characterization of Valid RoPE}
\label{sec:constraints of rope}
\vspace{-0.5em}
In this section, we present and prove the necessary and sufficient conditions for the $N$-dimensional RoPE to satisfy both \textit{relativity} and \textit{reversibility}, which serve as the foundational constraints for constructing any valid RoPE.
\begin{theorem}
\label{thm:constraints}
A set of generators \(\{B_1, \dots, B_N\} \subset \mathbb{R}^{d \times d}\) defines a valid N-dimensional RoPE 
\(R_{\boldsymbol{x}} = \exp\left(\sum_{i=1}^N x_i B_i\right)\) that satisfies both relativity and reversibility 
if and only if the following conditions hold:
\begin{enumerate}
    \item \(B_i \in \mathfrak{so}(d)\) for all \(i\),
    \item \([B_i, B_j] = 0\) for all \(i, j \in [N]\),
    \item \(\{B_1, \dots, B_N\}\) are linearly independent, 
    \item Any \(x_i\) must lie in the minimal period \(T_i\) of periodic \(B_i\), and \(x_i \in \mathbb{R}\), if \(B_i\) is aperiodic.

\end{enumerate}
\end{theorem}
\vspace{-0.5em}
Conditions 1–2 relate to the property of \textit{relativity}, while Conditions 3–4 concern \textit{reversibility}. Thus, we divide the analysis into two parts.

\vspace{-0.7em}
\begin{proof}
Let
\[
  A_1 = \sum_{i=1}^{N} u_i B_i,
  \qquad
  A_2 = \sum_{i=1}^{N} v_i B_i,
  \qquad
  (\boldsymbol{u}, \boldsymbol{v}) \equiv (\boldsymbol{x_1},\boldsymbol{x_2}) \in \mathbb{R}^{N}\times\mathbb{R}^{N}.
\]
\paragraph{Conditions 1 and 2}
To verify relativity, we analyze the matrix identity implied by \eqref{eq:nd relativity} and \eqref{eq:rope form}:
\begin{equation}
    \exp(A_1)^{\top}\,\exp(A_2)
    \;=\;
    \exp(A_2-A_1).
    \label{eq:rope_identity}
\end{equation}

\textbf{Sufficiency.}
If conditions 1 and 2 hold, then both \(A_1\) and \(A_2\) are skew-symmetric and commute. Therefore,
\(
  \exp(A_1)^{\top} = \exp(-A_1),
\)
and
\[
  \exp(-A_1)\exp(A_2)
    = \exp(-A_1 + A_2)
    = \exp(A_2 - A_1),
\]
which confirms identity \eqref{eq:rope_identity}, proving the sufficiency.

\textbf{Necessity.}
Define
\(
  F(\boldsymbol{u}, \boldsymbol{v}) = \exp(A_1)^{\top}\exp(A_2) - \exp(A_2 - A_1).
\)
We prove necessity by the low-order partial derivatives of \(F\).
Since \(F \equiv 0\) according to \eqref{eq:rope_identity}, all derivatives at the origin must vanish:
\[
  \partial_{u_i}F(0,0) = B_i^{\top} + B_i = 0
  \quad\Longrightarrow\quad B_i^{\top} = -B_i \quad \text{(Condition 1)},
\]
\[
  \partial_{v_j}F(0,0) = B_j - B_j = 0
  \quad\Longrightarrow\quad B_j = B_j, \quad \text{(trivially holds, no constraint)}.
\]
\[
  \partial_{u_i}\partial_{v_j}F(0,0)
    = -B_i B_j + B_j B_i = -[B_i, B_j] = 0
  \quad\Longrightarrow\quad [B_i, B_j] = 0 \quad \text{(Condition 2)}.
\]
These necessary conditions are derived from evaluating derivatives at the origin, but they hold over the full domain of \(\boldsymbol{u}, \boldsymbol{v}\) by the analyticity of \(F\). 
Since these conditions also ensure sufficiency, no stronger algebraic constraints can exist without violating sufficiency.
Thus, Conditions 1 and 2 are both necessary and sufficient. 
Full derivations, including an alternative approach using complete Taylor expansions without coordinate fixing, are provided in the Appendix~A.1.

\paragraph{Conditions 3 and 4} 
For reversibility, we analyze the matrix holding the following identities according to \eqref{eq:rope form} and \eqref{eq:nd reversibility}:
\[
    \exp (A_1) = \exp(A_2) \quad \Longrightarrow \quad \boldsymbol{u} = \boldsymbol{v},\forall u_i, v_i \in [0, T_i]
\]

\textbf{Sufficiency.}  
Assume \(\{B_i\}\) are linearly independent. Let \(\Delta_i = v_i - u_i\).  
Commutativity gives
\[
\exp\left(\sum_i \Delta_i B_i\right) = \prod_i \exp(\Delta_i B_i) = I.
\]
By Condition 4, if the generators are periodic, they share a common minimal period normalized to 1, so \(\Delta_i \in \mathbb{Z}\); if aperiodic, then \(\exp(\Delta_i B_i) = I\) implies \(\Delta_i = 0\).  
In both cases, Condition 3 linear independence forces all \(\Delta_i = 0\). Hence \(\boldsymbol{u} = \boldsymbol{v}\), proving the sufficiency.

\textbf{Necessity.} Injectivity forces
\[
\sum_i\Delta_i B_i = 0
\;\Longrightarrow\;
\exp\!\Bigl(\sum_i\Delta_i B_i\Bigr)=I
\;\Longrightarrow\;
\Delta=0,
\]
so \(\{B_i\}\) are linearly independent (Condition 3), otherwise a nonzero \(\Delta\) would break injectivity.  

If \(\exp(T_iB_i)=I\), then
\(
\exp(\dots,x_i+T_i,\dots)
=\exp(\dots,x_i,\dots),
\)
which would violate injectivity unless \(x_i\in[0,T_i)\); if \(B_i\) is aperiodic, then \(x_i\mapsto\exp(x_iB_i)\) is injective on \(\mathbb{R}\) (Condition 4).
\end{proof}
\vspace{-0.8em}
It is worth noting that \(B_i\) with irrational eigenvalue ratios are aperiodic and are not considered in the main text; their analysis is deferred to the Appendix A.2. Accordingly, Condition~4 ensures the frequency of each  \(B_i\) is sufficient to cover the full token range along its associated dimension, such as the text length in 1D.
In contrast, Conditions~1--3 impose algebraic constraints on the generator set \( \{B_i\} \), which guide its construction and constitute the main focus of this paper.

\vspace{-0.6em}
\section{Rethink Standard 1D and 2D RoPE}
\label{sec:rethink 1d and 2d RoPE}
\vspace{-0.6em}
In this section, we show that the standard 1D RoPE and 2D RoPE, are both special cases of the general formulation in \eqref{eq:rope form}, under the Conditions 1-4 of Theorem~\ref{thm:constraints}.
We further analyze the standard 2D RoPE, highlighting its computational efficiency and its limitation in lacking interactions across spatial dimensions.

\vspace{-0.6em}
\subsection{1D RoPE}
\label{subsec:1d RoPE}
For 1D case, as we only need to a single generator \(B \in \mathfrak{so}(n)\), the Conditions 1, 2 and 3 are naturally satisfied. Notably, in the simplest case where $n = 2$, a valid choice is:
\[
    B = \theta
    \begin{bmatrix}
    0 & -1 \\
    1 & 0
    \end{bmatrix}.
\]
As a result, its corresponding matrix exponential gives the standard 1D RoPE formulation:
\begin{equation}
    \label{eq:find 1d rope}
    R_m = \exp(mB) = 
    \begin{bmatrix}
    \cos(m\theta) & -\sin(m\theta) \\
    \sin(m\theta) & \cos(m\theta)
    \end{bmatrix}.
\end{equation}
where \(m\) is the absolute position of each token.
The Condition 4 requires that the absolute position of each token lies within a full rotation period of the generator, which is critical for preserving injectivity of the RoPE. In practice, most modern LLMs, such as \citep{llama3}, satisfy this condition by selecting rotation frequencies such that the period of each generator exceeds the model’s maximum sequence length. This guarantees that all token positions encountered during training and inference remain within a single period, thereby ensuring local injectivity in real-world applications.

\vspace{-0.4em}
\subsection{2D RoPE}
In the 2D case, to satisfy Conditions~1, 2, and 3, the generators \(B_1\) and \(B_2\) must meet the following criteria:
\(
    B_1, B_2 \in \mathfrak{so}(n), \text{ and } [B_1, B_2] = 0.
\)
A key structural property of the Lie algebra \(\mathfrak{so}(n)\) is that any skew-symmetric matrix can be block-diagonalized via an orthogonal matrix \(Q\) (see Appendix~A.3). 
Specifically, each matrix \(B \in \mathfrak{so}(n)\) admits the decomposition:
\(
    B = Q\, \mathrm{diag}(J_{\lambda_1}, J_{\lambda_2}, \dots, J_{\lambda_k})\, Q^\top,
\)
where each block \(J(\lambda_k)\) is a \(2 \times 2\) canonical skew-symmetric matrix of the form
\[
J(\lambda_k) =
\begin{bmatrix}
0 & -\lambda_k \\
\lambda_k & 0
\end{bmatrix},
\]
associated with a complex conjugate eigenvalue pair \(\pm i\lambda_k\).
When \(n\) is odd, this decomposition may also include a scalar zero block, corresponding to a zero eigenvalue. In this work, we focus on the even-dimensional case, which is common in practice since the head dimensions in Transformer models are typically even.

Simultaneously identifying a shared orthogonal matrix \(Q\) that block-diagonalizes both \(B_1\) and \(B_2\) can be nontrivial. To simplify implementation, we consider a construction where both \(B_1\) and \(B_2\) are directly defined as block-diagonal matrices. This corresponds to choosing \(Q = I\), resulting in:
\begin{equation}
    \label{eq:simplest B1 and B2}
    \begin{aligned}
    B_1 = \mathrm{diag}(J_{\lambda_1}, J_{\lambda_2}, \dots, J_{\lambda_k}), \quad
    B_2 = \mathrm{diag}(J_{\mu_1}, J_{\mu_2}, \dots, J_{\mu_k}),
    \end{aligned}
\end{equation}
where \(k\) is the number of \(2 \times 2\) rotational blocks. This construction guarantees that Conditions~1 and 2 are satisfied: each matrix is explicitly skew-symmetric and their block-diagonal structure ensures commutativity, thereby preserving relativity. A formal verification is provided in Appendix A.4.
However, \(B_1\) and \(B_2\) defined in this way may not be linearly independent, which violates Condition~3.

To ensure linear independence, one simple and effective strategy is to let \(B_1\) and \(B_2\) act on orthogonal \(2 \times 2\) subspaces. Specifically, we assign nonzero Jordan blocks in non-overlapping positions, setting the remaining blocks to zero. For example:
\begin{equation}
    \label{eq:2d B1B2}
    B_1 =
    \begin{bmatrix}
        J & 0 \\
        0 & 0
    \end{bmatrix}, \qquad
    B_2 =
    \begin{bmatrix}
        0 & 0 \\
        0 & J
    \end{bmatrix}, \quad \text{where} \quad
    J =
    \begin{bmatrix}
        0 & -1 \\
        1 & 0
    \end{bmatrix}.
\end{equation}

This leads to the standard 2D RoPE formulation as a special case:
\begin{equation}
    \label{eq:standard 2d rope}
    \begin{split}
        R_{(u_i, v_i)} &= \exp(u_i B_1 + v_i B_2) \\
        &= \mathrm{diag}\left( R(u_i), R(v_i) \right)
    \end{split}
    \quad \text{where } 
    R(t) = 
    \begin{bmatrix}
        \cos(t \theta) & -\sin(t \theta) \\
        \sin(t \theta) &  \cos(t \theta)
    \end{bmatrix}.
\end{equation}

This construction lies in \(\mathrm{SO}(4)\) and satisfies Conditions~1–3. Similar to the 1D case, Condition~4 is satisfied by choosing frequencies such that the periods of \(B_1\) and \(B_2\) cover the full token range along each spatial dimension, ensuring injectivity in practice.

\vspace{-0.5em}
\subsection{Discussion of Standard 2D RoPE}
\label{subsec:discussion of 2d rope}
It is important to note that the standard 2D RoPE formulation in Eq.~\eqref{eq:standard 2d rope} is a special case that satisfies the conditions of Theorem~\ref{thm:constraints}, where each spatial dimension is encoded by independent rotations on orthogonal planes. One key advantage of this design is its computational efficiency: each rotation can be computed using Euler’s formula in the complex domain, avoiding explicit trigonometric or matrix exponential operations (see Appendix A.5).

However, this separation also leads to a key limitation: the positional encoding across spatial dimensions is decoupled, which prevents the model from capturing cross-dimensional interactions.  
In effect, the standard RoPE behaves similarly to a Manhattan metric, emphasizing axis-aligned relationships while failing to represent diagonal or joint positional structures.

In summary, although the standard 2D RoPE satisfies Theorem~\ref{thm:constraints} and offers high computational efficiency, its lack of cross-dimensional interaction may limit its representational capacity. This motivates the exploration of more expressive designs that enable interactions across dimensions while preserving theoretical guarantees and computational efficiency. We discuss such constructions in detail in Section~\ref{N-dimensional RoPE construction via MASA}.

\vspace{-0.5em}
\section{N-dimensional RoPE}
\label{N-dimensional RoPE construction via MASA}
\vspace{-0.5em}
In this section, we present the main theoretical results with proof for constructing $N$-dimensional RoPE, showing that the valid forms correspond to a basis of a Maximal Abelian Subalgebra (MASA) in $\mathfrak{so}(2N)$. We then discuss how to introduce cross-dimensional interactions while preserving the essential RoPE properties and computational efficiency.

\vspace{-0.5em}
\subsection{RoPE Construction via MASA}
\label{subsec:masa construction}
\vspace{-0.5em}
To construct an $N$-dimensional RoPE satisfying the skew-symmetry, commutativity, and linear independence from Condition 1-3 in Theorem~\ref{thm:constraints}, we seek a set of generator matrices 
\(\{B_i\}\) forming an Abelian subalgebra of $\mathfrak{so}(d)$ that contains at least $N$ linearly independent generators. 
A natural candidate is a Maximal Abelian Subalgebra (MASA), a subalgebra of $\mathfrak{so}(d)$ that is Abelian and not properly contained in any larger Abelian subalgebra. The dimension of a MASA equals the rank of $\mathfrak{so}(d)$, which is $\left\lfloor \tfrac{d}{2} \right\rfloor$, and thus this quantity defines the upper bound on the number of commuting generators.
We now state three key theorems that characterize the space of valid RoPE constructions from this Lie-theoretic perspective:

\begin{theorem}
\label{the1}
$N$-dimensional RoPE can be realized as a basis of a MASA in $\mathfrak{so}(2N)$.
\end{theorem}

\begin{theorem}
\label{the2}
The standard $N$-dimensional RoPE corresponds to the maximal toral subalgebra in $\mathfrak{so}(2N)$.
\end{theorem}

\begin{theorem}
\label{the3}
For any $\left\lfloor \tfrac{d}{2} \right\rfloor > N$, the $N$-dimensional RoPE can be constructed by selecting $N$ linearly independent generators from a MASA of $\mathfrak{so}(d)$.
\end{theorem}

These results collectively imply that a necessary condition for the existence of such a RoPE is:
\(
\left\lfloor \frac{d}{2} \right\rfloor \geq N,
\)
ensuring that the ambient Lie algebra has sufficient degrees of freedom to accommodate $N$ commuting generators.
\begin{lemma}
A maximal Abelian subalgebra (MASA) $\mathfrak{a}$ satisfies:
\begin{enumerate}
    \item $[X, Y] = 0$ for all $X, Y \in \mathfrak{a}$ (Abelian)
    \item If another subalgebra $\mathfrak{a}'$ satisfies $\mathfrak{a} \subset \mathfrak{a}'$, then $\mathfrak{a}' = \mathfrak{a}$ (maximality)
\end{enumerate}
\end{lemma}

\vspace{-0.5em}
\paragraph{Case 1: \(\left\lfloor \tfrac{d}{2} \right\rfloor = N\)}  
In this case, selecting \(N\) generators from a MASA of \(\mathfrak{so}(d)\) amounts to choosing a full basis of the MASA. Such a basis guarantees mutual commutativity and linear independence, thereby fulfilling Conditions 1–3 in Theorem~\ref{thm:constraints} and proving Theorem~\ref{the1}.

A concrete realization of this construction is the \emph{maximal toral subalgebra}, which decomposes \(\mathbb{R}^{2N}\) into \(N\) orthogonal 2D planes. Each basis element corresponds to an independent planar rotation, represented by a non-overlapping \(2 \times 2\) skew-symmetric matrix. The matrix exponential of their weighted sum then yields the standard \(N\)-dimensional RoPE:
\begin{equation}
    \label{eq:standard nd rope}
    \begin{split}
        R_{\mathbf{x}} &= \exp\left( \sum_{i=1}^N x_i B_i \right) \\
                       &= \mathrm{diag}\big( R(x_1), R(x_2), \dots, R(x_N) \big),
    \end{split}
    \quad \text{where } 
    R(t) = 
    \begin{bmatrix}
        \cos(t\theta) & -\sin(t\theta) \\
        \sin(t\theta) &  \cos(t\theta)
    \end{bmatrix}.
\end{equation}
This proves Theorem~\ref{the2}.

\vspace{-0.5em}
\paragraph{Case 2: $\left\lfloor \tfrac{d}{2} \right\rfloor > N$}
In this case, since any MASA in \(\mathfrak{so}(d)\) contains exactly \(\left\lfloor \tfrac{d}{2} \right\rfloor\) mutually commuting generators, it is always possible to select \(N\) of them to construct a valid RoPE. This provides a verification of Theorem~\ref{the3}.
We include this case for completeness, although it is not the main focus of this study.

Regarding Condition~4, similar to the 1D and 2D cases, one can choose appropriate frequencies for each dimension to ensure that the resulting periods cover the full token range.

\vspace{-0.5em}
\subsection{Learning Dimensional Interactions}
\vspace{-0.5em}

In this paper, we primarily focus on \textbf{Case 1} in Section~\ref{subsec:masa construction}, where the core problem reduces to constructing a basis of the MASA, as stated in Theorem~\ref{the1}.  
A natural solution is the maximal toral subalgebra or standard \(N\)-dimensional RoPE in \eqref{eq:standard nd rope}, which is computationally efficient but encodes each spatial dimension independently, without capturing cross-dimensional interactions.

This raises two fundamental questions in constructing a MASA basis whose matrix exponential defines a valid high-dimensional RoPE:
\textbf{Q1}: How can we enable dimensional interactions?
\textbf{Q2}: How can we maintain computational efficiency?

\vspace{-0.7em}
\paragraph{Q1: Introducing dimensional interaction.} A straightforward solution is to directly learn a MASA basis from data instead of maximal toral subalgebra. However, this approach faces significant challenges: ensuring commutativity and linear independence throughout optimization is non-trivial, and naive parameterizations often incur substantial computational overhead.
To address \textbf{Q1}, we propose a change-of-basis formulation. Specifically, we begin by constructing a set of diagonal block matrices \(\{B_i\}\) that form a valid MASA basis, explicitly satisfying the constraints in Theorem~\ref{thm:constraints}. 
To model dimensional interactions, we make the orthogonal matrix $Q$ learnable, and define the RoPE transformation as:
\(
    R_{\boldsymbol{x}} = \exp(\textstyle\sum x_i Q B_i Q^\top) 
    = Q \exp(\textstyle\sum x_i B_i) Q^\top.
\)
The resulting transformation satisfies relativity by the identity:
\begin{align}
    R_{\boldsymbol{u}}^\top R_{\boldsymbol{v}} 
    &= (Q \exp(\textstyle\sum u_i B_i) Q^\top)^\top (Q \exp(\textstyle\sum v_i B_i) Q^\top) \notag \\
    &= Q \exp(\textstyle\sum u_i B_i) \exp(\textstyle\sum v_i B_i) Q^\top \notag \\
    &= Q \exp\left(\textstyle\sum (v_i - u_i) B_i\right) Q^\top \notag \\
    &= R(\boldsymbol{v} - \boldsymbol{u})
    \label{eq:relative_rope}
\end{align}
and  reversibility:
\(
    R_{\boldsymbol{u}}^\top R_{\boldsymbol{v}} 
    \Rightarrow \exp(\textstyle\sum u_i B_i) = \exp(\textstyle\sum v_i B_i)
    \Rightarrow \boldsymbol{u} = \boldsymbol{v}.
\)
\textbf{Q1} thus reduces to learning an orthogonal matrix \( Q \) to introduce inter-dimensional interactions, for which we consider the Cayley transform and Householder as candidate parameterizations.


\textbf{Cayley Transform.}
The Cayley transform \citep{optimization} maps a skew-symmetric matrix \( A \in \mathfrak{so}(n) \) to an orthogonal matrix \( Q \in \mathrm{SO}(n) \) via
\(
    Q = (I - A)(I + A)^{-1}.
\)
It provides a compact and efficient parameterization, requiring only the upper-triangular entries of \( A \), and guarantees orthogonality as long as \( I + A \) is invertible. By making \( A \) learnable, this formulation enables flexible modeling of interactions across spatial dimensions.
However, the Cayley transform only spans the  \( \mathrm{SO}(n) \), which consists rotations and excludes reflections, limiting its ability. 
More critically, without explicit control over $A$, the resulting transformation may introduce uncontrolled global interactions and disrupt the local rotational structure of RoPE, potentially undermining its effectiveness.

\textbf{Householder.}
In contrast to the Cayley transform, Householder transformations can represent the full orthogonal group \( \mathrm{O}(n) \), including both rotations and reflections. 
Each Householder matrix is defined as \( H = I - 2vv^\top \), where \( v \in \mathbb{R}^n \) is a unit vector. A general orthogonal matrix can be expressed as a product of \( k \) such reflections, allowing the construction of arbitrary orthogonal transforms. 
By learning the vectors \( \{v_i\}_{i=1}^k \), interactions between spatial dimensions can be introduced in a highly expressive yet structured way.
This formulation, although involving higher computational cost due to the composition of multiple reflections, provides stronger expressive power while keeping the local structures.
Our empirical results and structural analysis further validate the observed differences between the two parameterizations.

\vspace{-0.7em}
\paragraph{Q2: Maintaining Computational Efficiency.}
Given that the proposed construction satisfies relativity, reversibility, and allows inter-dimensional interactions via the orthogonal matrix \( Q \), the remaining objective is to ensure that \( \exp\left(\sum x_i B_i\right) \) remains computationally efficient.
To this end, the maximal toral subalgebra offers a particularly natural and elegant solution as shown in Theorem \ref{the2} with high computed efficiency using Euler’s formula (see Appendix A.6). The overall structure becomes:
\begin{equation}
    \label{eq:framawork}
    \exp\left(\sum x_i Q B_i Q^\top\right) = Q \cdot \bigoplus_{j=1}^{N} \exp\left( \sum_{i=1}^N x^{(i)} J_{i,j} \right) \cdot Q^\top,
\end{equation}
where \(i=1,\dots,N\) corresponds to spatial dimensions and \(j=1,\dots,N\) indexes the \(2\times2\) toral blocks on the diagonal. Each inner block computations \( \exp\left( \sum_{i=1}^N x^{(i)} J_{i,j} \right) \) is computed once before training and remains fixed thereafter. By computing the blockwise exponentials in advance and learning only the orthogonal matrix \( Q \), our formulation simultaneously solve \textbf{Q1} and \textbf{Q2}.
Other parameterizations, such as Givens rotations and matrix exponentials, are left to future work.
In practice, the \( Q \) and \( Q^\top \) cancel out in the transform \( Q \exp(\cdot) Q^\top \), as in \eqref{eq:relative_rope}, reducing it to a simple right multiplication by \( Q^\top \) after standard RoPE without architectural changes.

\vspace{-0.5em}
\section{Experiments}
\label{sec:experiments}
\vspace{-0.5em}
To evaluate the effectiveness of different types of RoPE, including the standard one as well as the Cayley and Householder variants, we conduct experiments on 2D visual inputs using two representative backbone models: Vision Transformer (ViT) \citep{vit} and Masked Autoencoder (MAE) \citep{mae}. We replace their original positional encoding with the respective RoPE variants while keeping all other training settings unchanged. Both ViT and MAE are used in their tiny configurations with approximately 22 million parameters. Our goal is not to pursue state-of-the-art task performance, but to provide a fair and controlled comparison across RoPE variants. To this end, we avoid hyperparameter tuning, focusing instead on validating the learning effectiveness of the orthogonal matrix $Q$.

\vspace{-0.6em}
\subsection{Supervised Classification}
\label{subsec:vit}
To evaluate different RoPE variants in supervised classification, we adopt a standard ViT-Tiny architecture without any distillation tricks or performance enhancements. All models are trained on ImageNet-Tiny \citep{tinyimagenet} for 150 epochs, and we report the best validation accuracy.
The detailed hyperparameters and training results are provided in the Appendix B.1.
As shown in Table~\ref{tab:comparison}, the Householder RoPE variant achieves the highest Top-1 accuracy among all methods. In contrast, the Cayley variant performs the worst, even below the standard RoPE. This suggests that not all inter-dimensional interaction mechanisms are beneficial, a point we further analyze in the next section.
However, the performance gain is relatively small, which we attribute to the fact that classification tasks based on ViT are generally not highly sensitive to positional encoding. As a result, the type of RoPE used and the inclusion of cross-dimensional interactions have limited impact in this setting.

\begin{table}[htbp]
\vspace{-0.5em}  
\centering
\begin{tabular}{cccc}
\hline
RoPE Variants & ViT Top-1 (\%) & MAE (Cls.) Top-1 (\%) & MAE (Seg.) mIoU (\%) \\ \hline
Standard      & 60.16     & 78.08                    & 36.71         \\
Cayley        & 59.97     & 78.08                    & 36.71         \\
Householder   & \textbf{60.39} & \textbf{78.13}       & \textbf{37.67} \\ \hline
\end{tabular}
\caption{Comparison of different RoPE variants on classification and segmentation tasks.}
\label{tab:comparison}
\vspace{-1.0em} 
\end{table}

\vspace{-0.8em} 
\subsection{Self-Supervised Training and Fine-tuning}
Since the reconstruction objective in MAE depends heavily on modeling spatial structure, it serves as an ideal testbed for evaluating positional encoding methods. 
Motivated by this, we adopt MAE as our framework and perform self-supervised pretraining on ImageNet-1K \citep{imagenet} for 400 epochs. 
Appendix~B.2 provides detailed hyperparameters and training results.
During pretraining, we replace the positional encoding of encoder with different RoPE variants, while keeping the decoder unchanged. 
We then fine-tune the pretrained models on both image classification and semantic segmentation tasks to comprehensively assess the effectiveness of each RoPE variant.

\vspace{-0.8em}
\paragraph{Classification fine-tuning.} To evaluate the performance of different RoPE variants after pretraining, we perform full classification fine-tuning on MAE for 100 epochs and report the best validation accuracy. Training details and results are provided in Appendix~B.3.
As shown in the Table \ref{tab:comparison}, the results are consistent with those observed in supervised ViT classification: the Householder variant achieves the highest Top-1 accuracy, while Cayley and the standard RoPE perform identically. 
The overall differences remain small, indicating that dimensional interactions introduced by RoPE have limited impact on classification tasks.

\vspace{-0.5em}
\paragraph{Segmentation fine-tuning.} To assess the impact of different RoPE variants in a position-sensitive task, we evaluate MAE on the ADE20K \citep{ade20k} semantic segmentation benchmark.
Specifically, we adopt the UPerNet framework \citep{upernet} and replace its backbone with our pretrained MAE models using different RoPE variants. Following the standard setting \citep{visionllama}, we perform full model fine-tuning for 160K iterations and report the best validation mIoU.
See Appendix~B.4 for hyperparameter settings and training results.
As shown in Table~\ref{tab:comparison}, unlike the classification task, the performance gap becomes more pronounced in segmentation.
Householder achieves the highest mIoU, while Cayley and standard RoPE yield identical results.
This aligns with our hypothesis that not all interaction patterns are equally suitable for position-sensitive tasks. While Cayley restricts \( Q \) to \(SO(n)\) via skew-symmetric parameterization, it may introduce irregular global interactions that disrupt local structure—an essential aspect in segmentation. A more detailed analysis of this effect is provided in the next section.

\vspace{-0.6em}
\section{Structural Analysis of Learnable RoPE Variants}
\label{sec:analysis}
\vspace{-0.6em}
In above theoretical analysis, RoPE encodes $N$-dimensional positions in $2N$ dimensional representation, where each spatial axis corresponds to a single 2x2 rotation block.
In practice, however, Transformer embeddings are typically much higher dimensional than the minimal $2N$, and each spatial axis is associated with multiple rotation blocks at different frequencies.
This design gives rise to two desirable properties: (1) preserving the rotational structure within each spatial-frequency block, and (2) enabling interaction across spatial dimensions.

\begin{figure}[htbp]
\vspace{-0.6em}
    \centering
    \begin{minipage}[t]{0.48\linewidth}
        \centering
        \includegraphics[width=\linewidth]{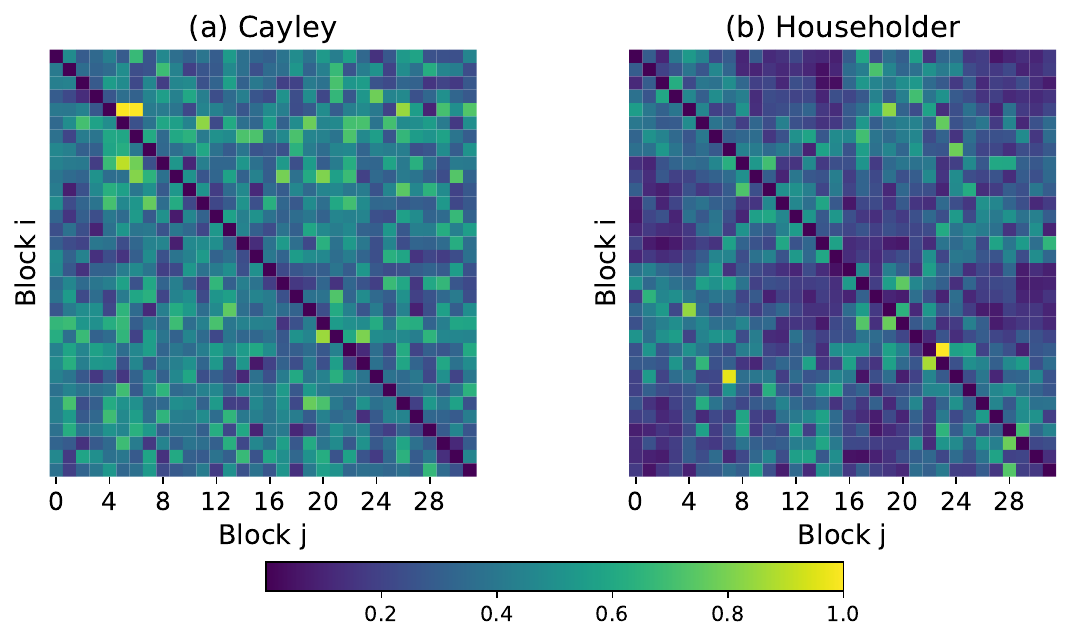}
        \caption{Relative structure deviation heatmaps.}
        \label{fig:structure_change}
    \end{minipage}%
    \hfill
    \begin{minipage}[t]{0.43\linewidth}
        \centering
        \includegraphics[width=\linewidth]{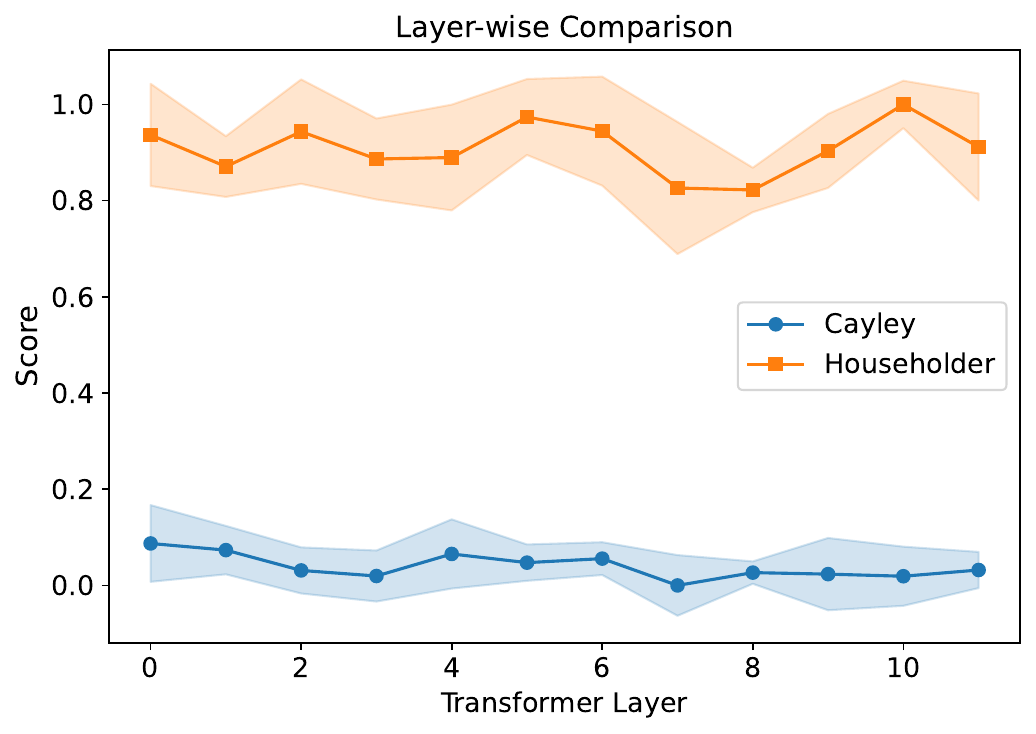}
        \caption{Layer-wise scores.}
        \label{fig:structure_interaction}
    \end{minipage}
\vspace{-0.6em}
\end{figure}

To analyze how well these properties are preserved in different learnable RoPE variants and understand their performance differences, we extract the trained \( Q \) in MAE constructed via Cayley and Householder parameterizations, and use them to compute the final learned RoPE via \eqref{eq:framawork}. The learned RoPE matrices are first averaged over all tokens, heads, and layers to obtain a global structure. 
Then, to reveal localized structural changes, we compute the relative Frobenius norm differences over 2×2 blocks with respect to the standard RoPE.
Figure~\ref{fig:structure_change} visualizes the structural deviations. In (a), the Cayley-based representation exhibits globally irregular cross-block interactions and fails to preserve local coherence. In contrast, (b) shows that Householder transformation better maintains local structure while enabling more organized inter-block interactions. These patterns suggest that the performance gap may be attributed to Cayley's tendency to disrupt locality.

To quantitatively compare Cayley and Householder in terms of structural behavior, we compute a score that reflects the balance between local structure preservation and global inter-block interactions. Specifically, we measure the relative Frobenius norm change within each \(2 \times 2\) rotation block to quantify local distortion, and the energy outside these blocks, corresponding to rotations in the other spatial dimension, to capture cross-dimensional interactions.
The score is defined as the ratio of inter-block energy to local distortion, with higher values indicating better structural balance. 
As shown in Figure~\ref{fig:structure_interaction}, Householder consistently achieves higher scores than Cayley, suggesting a more favorable trade-off between locality and inter-dimensional interaction.
Additional analysis details and results for both MAE and ViT are provided in Appendix~C.


\vspace{-1em}
\section{Conclusion and Limitations}
\label{sec:conclusion}
\vspace{-0.6em}
Overall, this work presents a principled theoretical framework for RoPE grounded in Lie group and Lie algebra theory. The necessary and sufficient conditions are derived for constructing valid RoPEs, which correspond algebraically to identifying a basis of the MASA in \(\mathfrak{so}(n)\). This perspective not only recovers the standard RoPE as a special case, but also enables structured extensions via learnable orthogonal transformations. Moreover, our experiments and analysis of the Cayley and Householder parameterizations suggest that the design of learnable RoPEs for dimensional interactions should carefully balance local structure preservation with inter-dimensional flexibility. We believe that this work lays a foundation for understanding RoPE and inspires the development of more robust and interpretable positional encoding strategies.

The main limitation of this study lies in the lack of large-scale experiments across a wider range of tasks and higher-dimensional data such as video or robotics, which prevents drawing more generalizable conclusions about different RoPE variants. Additionally, further strategies for introducing inter-dimensional interactions while preserving local structure remain to be explored. We leave these directions for future work.

\bibliographystyle{plainnat}
\bibliography{reference}

\end{document}